\documentclass[11pt,a4paper]{article}
\usepackage[hyperref]{emnlp2020}
\usepackage{times}
\usepackage{latexsym}

\usepackage{cleveref}

\usepackage{todonotes}

\usepackage{microtype}

\definecolor{myred}{rgb}{0.73, 0.09, 0.09}

\definecolor{mypurple}{rgb}{0.3, 0, 0.6}

\definecolor{mygreen}{rgb}{0, 0.4, 0.11}

\aclfinalcopy
\def\aclpaperid{259}

\title{How well does surprisal explain N400 amplitude under different experimental conditions?}

\author{James A. Michaelov \\
  Deparmtment of Cognitive Science \\
  University of California, San Diego \\
  \texttt{j1michae@ucsd.edu} \\\And
  Benjamin K. Bergen \\
  Deparmtment of Cognitive Science \\
  University of California, San Diego \\
  \texttt{bkbergen@ucsd.edu} \\}

\date{}

\begin{document}
\maketitle
\begin{abstract}
We investigate the extent to which word surprisal can be used to predict a neural measure of human language processing difficulty---the N400. To do this, we use recurrent neural networks to calculate the surprisal of stimuli from previously published neurolinguistic studies of the N400. We find that surprisal can predict N400 amplitude in a wide range of cases, and the cases where it cannot do so provide valuable insight into the neurocognitive processes underlying the response.
\end{abstract}

\section{Introduction}
The N400 component of the event-related brain potential is generally understood to be a neural signal of processing difficulty \citep{kutas_thirty_2011}. After over 1,000 articles published on the topic, we know that all else being equal, an upcoming word that is supported by the semantics of the context will elicit a lower-amplitude N400 than a word that is not \citep{kutas_thirty_2011,kuperberg2020tale}. However, despite the great amount of experimental research on the topic, many aspects of the N400 are still not well understood.

In addition to `long-standing and recent linguistic [...] inputs' \citep[p.~641]{kutas_thirty_2011}, the context that impacts N400 amplitude is thought to include factors such as world experience, attentional state, and mood \citep{kutas_thirty_2011}. Over the last decade, there have been a number of attempts to use computational modeling to test hypotheses about the neurocognitive processes underlying the N400 and how the aforementioned factors may impact its amplitude \citep{parviz2011using,laszlo2012neurally,laszlo2014psps,rabovsky_simulating_2014,frank_erp_2015,ettinger_modeling_2016,cheyette2017modeling,brouwer2017neurocomputational,delaney2017comprehenders,rabovsky_modelling_2018,venhuizen_expectation-based_2018,fitz_sentence_level_2019}.

As the majority of experimental research on the N400 involves manipulating the relationship between the stimulus and the preceding linguistic context \citep{kutas_thirty_2011}, a computational account of how linguistic inputs impact N400 amplitude is a logical starting point. Language models are inherently models of linguistic prediction based only on language input. Since N400 amplitude reflects how unexpected an upcoming word is based on context, the predictions of a language model can be used to model how expected a word is based on the linguistic input, and thereby investigate the extent to which N400 amplitude is explainable by linguistic input alone. 

Recent research has shown that \textit{surprisal}, a measure of how unlikely a language model predicts the next word in sequence to be, correlates overall with N400 amplitude \citep{frank_erp_2015,aurnhammer2019evaluating}. Thus, to investigate the extent to which N400 amplitude is explained by linguistic input alone, we ask to what extent surprisal can explain the variance observed in N400 amplitude.

In order to investigate this, we run experimental stimuli from eleven experiments from six papers \citep{urbach_quantifiers_2010,kutas1993company,ito2016predicting,osterhout1995event,ainsworth1998dissociating,kim_independence_2005} through two recurrent neural network language models \citep{jozefowicz_exploring_2016,gulordava_colorless_2018}, systematically comparing the significant predictors of N400 amplitude and surprisal. We find that in the majority of cases, significant differences in surprisal predict significant differences in N400 amplitude, and discuss the implications of the cases where it does not.

\section{Background}
\subsection{The N400}\label{ssec:N400}
The N400 is a negative deflection in the event-related brain potential (ERP) that peaks roughly 400ms after the presentation of a stimulus \citep{kutas_reading_1980,kutas_thirty_2011}. Most current accounts agree that N400 amplitude reflects processing difficulty for a specific lexical item, where a lower amplitude reflects prior activation of some of the semantic content associated with the word \citep{kutas_thirty_2011,kuperberg_separate_2016,kuperberg2020tale}.

Recent research has found that N400 amplitude `\textit{decreases} with supportive context, but does not \textit{increase} when predictions are violated' (\citealp[p.~2]{delong2020comprehending}, emphasis in original; see \citealp{kutas_thirty_2011,van_petten_prediction_2012,luke_limits_2016,kuperberg2020tale}, for discussion). Crucially, therefore, we should not think of N400 amplitude as a general measure of prediction error. It is not the case that the N400 elicited by a word increases when the word is more semantically anomalous or unexpected based on the preceding context; rather, it is the case that N400 amplitude is reduced when the word is semantically congruous or predictable because it is facilitated by the preceding context. 

This facilitation can occur in a large number of ways. All else being equal, words that are more semantically congruous, typical, or plausible completions of a sentence elicit lower N400 amplitudes than words that are more semantically incongruous, atypical, and implausible completions, respectively \citep[e.g.][]{kutas_reading_1980,urbach_quantifiers_2010,ito2016predicting,osterhout1995event,ainsworth1998dissociating,kim_independence_2005,kutas_thirty_2011}.

One well-known correlate of N400 amplitude is the cloze probability \citep{taylor1953cloze,bloom1980completion} of a word---the probability that it will be offered to fill a specific gap in a sentence by a given sample of individuals in a norming study. All else being equal, higher-cloze completions elicit lower N400 amplitudes \citep{kutas_brain_1984,kutas_thirty_2011}. Additionally, even when matched for cloze, words semantically related to the highest-cloze completion elicit lower-amplitude N400s than unrelated words \citep{kutas1993company,federmeier_rose_1999,ito2016predicting}.

\subsection{Cognitive Plausibility of RNN-LMs in N400 modeling}
To disentangle the effect of linguistic input from other factors affecting N400 amplitude, a valid model of such linguistic input is needed. Recurrent Neural Network Language Models (RNN-LMs) are, in many ways, perfect models of the `long-standing and recent linguistic [...] inputs'  \citep[p.~641]{kutas_thirty_2011} thought to impact N400 amplitude. Long-standing linguistic inputs in humans are made up of previous language experience, which is analogous to a model's training data; and recent linguistic input is the linguistic context that impacts how humans understand the current utterance, which is analogous to the word sequence preceding the word to be predicted in the model's test data. 

Beyond being largely developed as models of human language comprehension \citep{elman1990finding}, recurrent neural network language models (RNN-LMs) have certain properties that make them reasonable models of human cognition. \citet{keller-2010-cognitively} identifies five features of the human language processing system that he argues are vital for a language model to be cognitively plausible. Three of these are exemplified by unidirectional RNN-LMs---like humans, they can make \textit{predictions} about upcoming words, have a distance-based \textit{memory cost}, and process language word-by-word in order in an  \textit{incremental} fashion (unlike bidirectional RNN-LMs and most transformer networks). The two remaining features, \textit{efficiency and robustness} and \textit{broad coverage} are determined more by the model's specific architecture and training than general architecture.

\subsection{Surprisal and N400 amplitude}
As discussed in \Cref{ssec:N400}, the neurolinguistic evidence suggests that the N400 is a measure of lexical processing difficulty. Recent work, both theoretical and experimental \citep[e.g.][]{hale2001probabilistic,levy2008expectation,boston2008parsing,demberg2008data,smith2008optimal,roark2009deriving,brouwer2010modeling,mitchell2010syntactic,monsalve2012lexical,fossum2012sequential,frank2012early,smith2013effect,frank2014modelling,willems2016prediction,delaney2017comprehenders}, has argued that surprisal, the negative  logarithm of the probability of a word $w_i$ given its preceding context $w_1...w_{i-1}$, as shown in \Cref{eq:surprisal}, is a good predictor of lexical processing difficulty.

\begin{equation}
    S(w_{i}) = -\log P(w_{i}|w_{1}...w_{i-1})
    \label{eq:surprisal}
\end{equation} 

Several researchers \citep{frank_erp_2015,delaney2017comprehenders,aurnhammer2019evaluating} have directly demonstrated that surprisal is correlated with N400 amplitude. In their study, \citet{delaney2017comprehenders} use a Bayesian approach to calculate the surprisal associated with a target word given a related or unrelated prime (using word association norms and word frequency), and find that this is correlated with N400 amplitude. \citet{frank_erp_2015} and \citet{aurnhammer2019evaluating} used a number of language models (including RNN-LMs) to calculate the surprisal of words in a natural language text, and compared this to the N400 elicited by these words in human participants, finding a statistically significant correlation.

\citet{frank_erp_2015} and \citet{aurnhammer2019evaluating} also find that surprisal is a better predictor of N400 amplitude than a number of RNN-LM-derived metrics based on the full probability distributions predicted by the model such as entropy. We suggest that this may be explained by the aforementioned finding that while the N400 amplitude for a word decreases when its semantic content has been pre-activated, it does not increase when a specific prediction is violated. In other words, N400 amplitude is a kind of positive prediction error---a measure of how not-predicted the target word was. This is what surprisal is by definition---it only takes into account how much the actual target word was predicted and is not affected by the rest of the probability distribution. The other metrics, on the other hand, also take into account the rest of the predicted probability distribution, which does not appear to be reflected in N400 amplitude. Thus, there is a theoretical reason for using surprisal to predict N400 amplitude based on previous neurolinguistics research.

\subsection{Predicting N400 effects}
An alternative approach, that taken by \citet{ettinger_modeling_2016}, is to use a language-model-derived metric as an analogue of the N400 and investigate whether experimental manipulations in the stimuli that result in statistically significant differences in N400 amplitude also result in statistically significant differences in the chosen metric. This approach allows researchers to investigate whether the reason for the correlation between the metric and N400 amplitude is in fact the experimental manipulation or some other factor. 

This is the general approach that we take in this study; however, rather than focusing on the cosine similarity between the word embedding of target word and the combined embeddings of the previous words in the sentence \citep{ettinger_modeling_2016}, we model N400 amplitude as surprisal \citep[following][]{frank_erp_2015,delaney2017comprehenders,aurnhammer2019evaluating}. Additionally, whereas \citeauthor{ettinger_modeling_2016}'s \citeyearpar{ettinger_modeling_2016} proof-of-concept paper is based on 40 sample sentences from a single study investigating one phenomenon, we use stimuli from eleven experiments (with over 100 sentences each) covering a wide range of phenomena.

\subsection{Other Models of N400 amplitude}
While a number of other researchers have used neural networks to model specific N400 findings this way \citep{laszlo2012neurally,laszlo2014psps,rabovsky_simulating_2014,cheyette2017modeling,brouwer2017neurocomputational,rabovsky_modelling_2018,venhuizen_expectation-based_2018,fitz_sentence_level_2019}, these studies differ in that these models all have semantic representations as part of their input or are trained to learn to output some form of semantic representation. Thus, these models are also limited to the experiments for which they were trained. 

For the same reason, these models can also not be used on their own to disentangle the effects of linguistic input from the semantic knowledge provided to them---this can only be done by comparison to models without this. While two of the studies compare their models to simple recurrent networks (SRNs) trained on the same data \citep{rabovsky_modelling_2018,fitz_sentence_level_2019}, these SRNs are not representations of the extent of what is possible with lingusitic input alone---these models are simple (for example, they do not use long short-term memory), and much of the power of RNNs comes from large training datasets \citep[see, e.g., the discussion in][]{chelba2013one}.

Finally, it should be noted that while all of the studies discussed in this section aim to model real N400 effects, only two \citep{laszlo2014psps,rabovsky_simulating_2014} use stimuli from real N400 experiments; in the remaining studies, stimuli are chosen to represent manipulations that studies have found to influence N400 amplitude. Given that the N400 is still not fully understood, it is important to verify that the experimental manipulations investigated actually do elicit the expected N400 effect. For this reason, we only use experimental stimuli provided for published N400 experiments, and compare the effect on surprisal directly to the reported effects on N400 amplitude.

\section{Approach, Motivations, and Hypotheses}
The aim of this study is to investigate the boundary conditions of using surprisal to model N400 amplitude. While there is evidence that surprisal and N400 amplitude are correlated overall \citep{frank_erp_2015,aurnhammer2019evaluating}, it is unclear what variance in N400 amplitude is actually being explained by surprisal. While it is tempting to assume that surprisal is correlated with the N400 because the same factors that lead to reduced N400 amplitudes lead to reduced surprisal, this has thus far not been shown empirically. 

This is the question that we investigate in this paper: which experimental manipulations that elicit a difference in N400 amplitude elicit the same difference in surprisal, and which do not?

We do this by running the (English language) stimuli from previously published N400 studies through two neural networks that have been used extensively to model human language processing \citep[e.g., in][]{wilcox_what_2018,futrell-etal-2019-neural,wilcox_structural_2019,an_representation_2019,costa_assessing_2020}. The two models used are the the best English LSTM from \citet{gulordava_colorless_2018} and BIG LSTM+CNN \textsc{inputs} from \citet{jozefowicz_exploring_2016}, henceforth \citep[following][]{futrell-etal-2019-neural} GRNN and JRNN, respectively. These models are both LSTM-RNN-LMs, but differ most notably in size and training data: The JRNN has two hidden layers (8192 and 1024 units), a 793471-word vocabulary, and was trained on 1 billion tokens \citep{chelba2013one}; while the GRNN has two hidden layers (both 650 units), a 50000-word vocabulary, and was trained on 90 million tokens.

In addition to answering questions about the nature of the neurocognitive systems underlying the N400, the results of this study also serve as a baseline for future research---they represent the best that current cognitively plausible neural network language models can do at predicting N400 amplitude using surprisal. Thus, future research that argues for additional sources of information or neurocognitive processes being involved in the N400 on the basis of modeling success should demonstrate that the inclusion of such components in the model improves upon the results presented here.

This aim of establishing a useful baseline is another reason for our choice of models---both are provided pre-trained by the authors, allowing for our results to be replicated and expanded upon. We also only use sets of stimuli that have been made available in papers or their supplementary materials. The stimuli from these papers \citep{urbach_quantifiers_2010,kutas1993company,ito2016predicting,osterhout1995event,ainsworth1998dissociating,kim_independence_2005}, which cover a range of experimental manipulations that are discussed in \Cref{sec:experiments}, are included in text format in our supplementary materials\footnote{\url{https://github.com/jmichaelov/does-surprisal-explain-n400}}.

\section{Experiments}\label{sec:experiments}

\begin{figure*}[!ht]
    \centering
    \includegraphics[width=\textwidth]{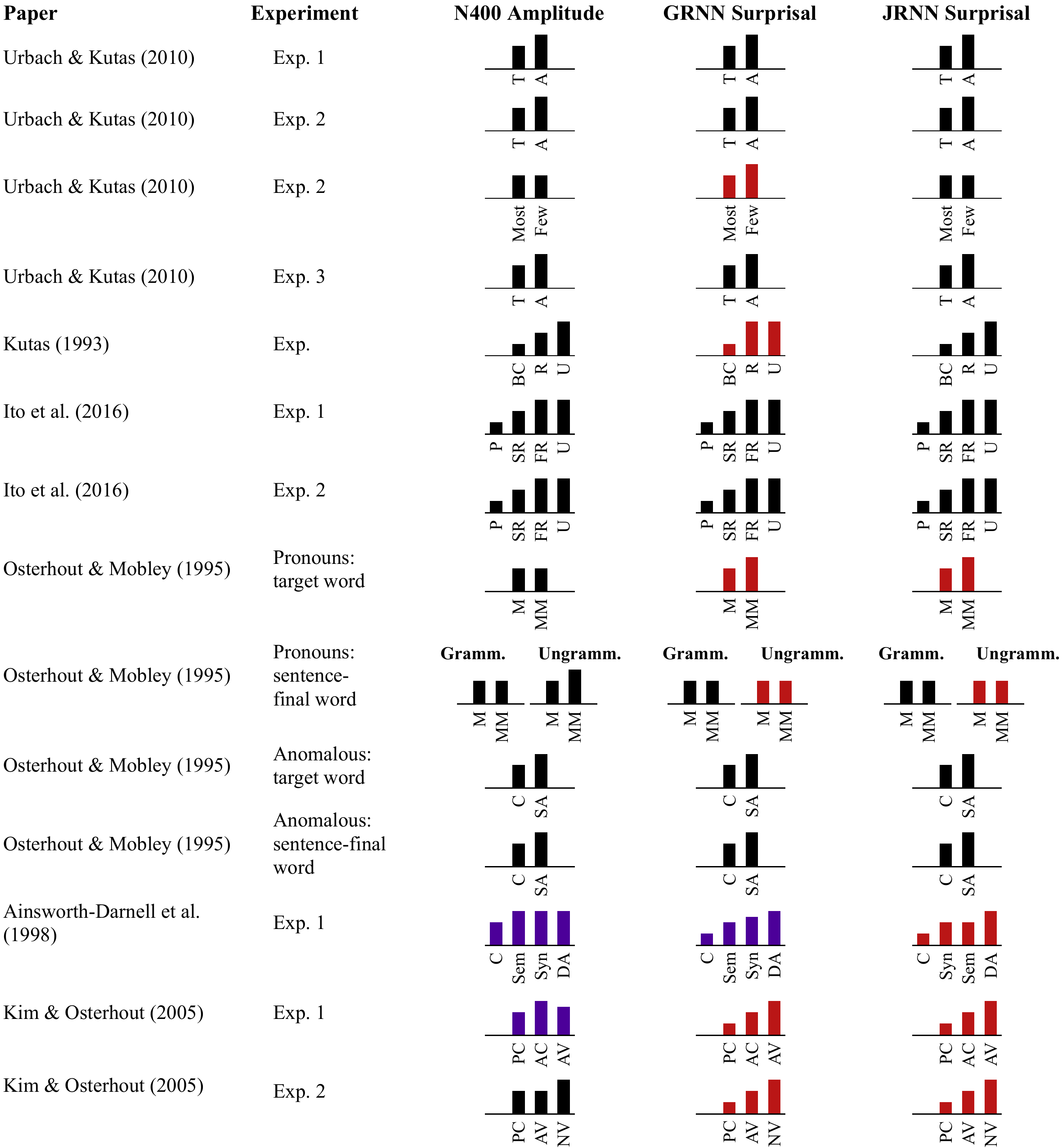}
    \caption{The significant differences between all conditions of significant predictors of N400 amplitude in the original studies and the surprisal of the GRNN and JRNN models. \textbf{Black} bars indicate successful modeling of the differences in N400 amplitude, \textcolor{myred}{\textbf{red}} bars indicate unsuccessful or partially unsuccessful modeling, and \textcolor{mypurple}{\textbf{purple}} bars indicate that the results are more complex than shown.}
    \label{fig:results}
\end{figure*}

\Cref{fig:results} is a visualization of the findings of the original N400 studies and the results of the simulations. Given the differences in measurements, there is no scale---the heights of the bars indicate which conditions elicited higher or lower N400 amplitudes or surprisals relative to the others in the same experiment or simulation. All and only the significant differences between conditions for significant predictors of the N400 or surprisal are shown, not including significant interactions with recording locations on the scalp (which are beyond the scope of the present study). Black bars represent successful modeling of the differences in N400 amplitude, red bars represent unsuccessful or partially unsuccessful modeling, and purple bars indicate that the results are more complex than can be represented in this way. Only stimuli sets with over 100 stimulus sentences were run through the models (GRNN and JRNN); and while the models were not able to predict the surprisal of all target words (due to limited vocabularies or being unable to process certain characters in sentences), both models successfully calculated the surprisals of over 100 target words in each study. Stimuli, target word surprisals, and the code used to run the models are all included in our supplementary materials.

Where possible, the significant predictors of the surprisal of the GRNN and JRNN models were selected via backwards model selection using likelihood ratio tests of linear-mixed effects models \citep{r,lme4} with and without the predictor under investigation as a main effect. When this was not possible, the significance of predictors were evaluated using a Type III ANOVA with Satterthwaite's method for estimating degrees of freedom \citep{kuznetsova2017lmertest}. Significant differences between experimental conditions (i.e. between the levels of a predictor) were calculated via t-test based on the selected linear-mixed effects model, using Satterthwaite's method to estimate degrees of freedom \citep{kuznetsova2017lmertest}. In this paper, significant predictors and significant differences between conditions are considered those where $p<0.05$ in the relevant statistical test. All code for the statistical analyses is included in our supplementary materials.

The remainder of this section discusses the experiments (and the original N400 studies on which they are based) in more detail.

\subsection{Urbach and Kutas (2010): Experiment 1}
Experiment 1 of \citet{urbach_quantifiers_2010} investigates the N400's sensitivity to the typicality of a patient of a described event. There were two kinds of sentences in this experiment exemplified by the following stimulus pair: \textit{prosecutors accuse \textbf{defendants}} (\textsc{typical}; T in \Cref{fig:results}) / \textit{\textbf{sheriffs}} (\textsc{atypical}; A) \textit{of committing a crime}. As expected, the N400 elicited by \textsc{typical} object nouns is significantly lower in amplitude than that elicited by \textsc{atypical} object nouns.

Typicality was also a significant predictor of the surprisal of both the GRNN and JRNN models (GRNN: $p<0.001$; JRNN: $p<0.001$), with \textsc{typical} object nouns eliciting a lower surprisal than \textsc{atypical} ones (GRNN: $p<0.001$; JRNN: $p<0.001$).

\subsection{Urbach and Kutas (2010): Experiment 2}
Expanding on Experiment 1, \citet{urbach_quantifiers_2010} ask whether the results are affected by whether the sentences begin with the word \textit{most} or \textit{few} (or synonymous expressions), e.g. \textit{\textbf{most} prosecutors accuse defendants}. The main effect of typicality remained. In addition, while the main effect of quantifier type was not significant overall (nor was there an interaction with typicality without an interacting electrode location variable), \citet{urbach_quantifiers_2010} found that \textsc{few}-type quantifiers reduced the N400 amplitude of \textsc{atypical} patients and reduced the extent to which N400 amplitude was lowered for \textsc{typical} patients, with this latter effect being found to be statistically significant via t-test. 

Typicality predicted the surprisals of both RNNs in the same direction as in Experiemnt 1 ($p<0.001$ for all statistical tests). The surprisal of the GRNN was also significantly predicted by quantifier type  ($p<0.001$), with \textsc{few}-type quantifiers eliciting significantly higher surprisals ($p<0.001$). As this pattern is limited only to the GRNN (and the analogous main effect does not appear in Experiment 3 for either model), this finding is not considered further. The t-test comparing the N400 of \textsc{typical} objects under the \textsc{few} and \textsc{most} quantifiers does not replicate with surprisal---there is no significant difference (GRNN: $p=0.107$; JRNN: $p=0.249$).

\subsection{Urbach and Kutas (2010): Experiment 3}
Experiment 3 of \citet{urbach_quantifiers_2010} is a variant of Experiment 2. Instead of \textsc{most} or \textsc{few} sentence beginnings, the words \textit{often} or \textit{rarely} appear after the subject (agent) noun, e.g. \textit{prosecutors \textbf{often} accuse defendants of committing a crime}. The aim of this was to investigate whether proximity of the quantifier to the target noun had an effect. \citet{urbach_quantifiers_2010} again found the same result---only typicality was a significant predictor of N400 amplitude overall; and a t-test found that the N400 reduction for \textsc{typical} nouns was attenuated by the word \textit{rarely}. 

GRNN and JRNN surprisals were only significantly predicted by typicality, with typical nouns eliciting a lower surprisal than atypical nouns ($p<0.001$ for all tests). The t-test comparing the N400 of \textsc{typical} objects under the \textsc{few} and \textsc{most} quantifiers does not replicate with surprisal---there is no significant difference (GRNN: $p=0.367$; JRNN: $p=0.283$).

\subsection{Kutas (1993)}
\citet{kutas1993company} examines the effect of relatedness to the \textsc{best completion} (the highest-cloze completion). An example of a \textsc{best completion} (BC) and \textsc{related} completion can be demonstrated by the following stimulus pair: \textit{The pizza was too hot to \textbf{chew}} (\textsc{related}; R) / \textit{\textbf{eat}} (BC). An example of a BC and \textsc{unrelated} pair is the following sentence: \textit{The paint turned out to be the wrong \textbf{consistency}} (\textsc{unrelated}; U)/ \textit{\textbf{color}} (BC). BC nouns were found to elicit the lowest N400 amplitude, followed by \textsc{related} nouns, followed by \textsc{unrelated} nouns. 
Experimental condition is a significant predictor of both GRNN and JRNN surprisal. However, while the surprisals in the GRNN are different between the BC and other nouns ($p<0.001$ for both \textsc{related} and \textsc{unrelated}), there is no significant difference between \textsc{related} and \textsc{unrelated} ($p=0.820$). On the other hand, the surprisals of the JRNN are lowest for BC nouns, followed by \textsc{related} nouns, followed by \textsc{unrelated} nouns ($p<0.001$ for all pairwise comparisons).

\subsection{Ito et al. (2016): Experiments 1 and 2}
\citet{ito2016predicting} further investigate the relatedness effect by investigating whether a word that is related in form to the most \textsc{predictable} word (i.e. the best completion) has a similar effect on N400 amplitude as being semantically related. The conditions can be illustrated with the following example sentence:\textit{ The student is going to the library to borrow a \textbf{book}} (\textsc{predictable}; P)/ \textit{\textbf{hook}} (\textsc{form-related}; FR)/ \textit{\textbf{page}} (\textsc{semantically related}; SR)/ \textit{\textbf{sofa}} (\textsc{unrelated}; U) \textit{tomorrow}. In both Experiments 1 and 2, where the difference was in the amount of time that the stimuli were presented, \citet{ito2016predicting} found that experimental condition was a significant predictor, and specifically that \textsc{predictable} words elicited the lowest N400 amplitude, followed by \textsc{semantically related} words, followed by the \textsc{form-related} and \textsc{unrelated} completions, which did not differ in N400 amplitude. 

We found the same pattern in the surprisal of both models ($p<0.001$ for condition as a predictor; $p<0.001$ for all significant pairwise comparisons; FR vs. U with GRNN surprisal: $p=0.080$; FR vs. U with JRNN surprisal: $p=0.399$).

\subsection{Osterhout and Mobley: Experiment 2}
\subsubsection{Pronoun Matching}
\citet{osterhout1995event} investigate the effect on the amplitude of the N400 elicited by words in sentences where pronouns either do or do not match a preceding noun, as illustrated in the following example: \textit{The aunt heard that \textbf{she}} (\textsc{match}; M) / \textit{\textbf{he}} (\textsc{mismatch}; MM) \textit{had won the lottery}. The \textsc{mismatch} sentences can be interpreted as grammatical sentences where the pronoun refers to a different person than that denoted by the sentence subject; or ungrammatical sentences, where the pronoun refers back to the sentence subject with the wrong gender. \citet{osterhout1995event} ask whether there is a difference in N400 amplitude between the two conditions, and whether this is affected by which interpretation is taken by participants.

\paragraph{Target Words}
First, \citet{osterhout1995event} look at the N400 measured at the pronoun itself, finding no significant effect of condition.

For both RNN-LMs, however, experimental condition is a significant predictor of surprisal, with matched pronouns eliciting a significantly lower surprisal ($p<0.001$ for all tests).

\paragraph{Sentence-Final Words}
The N400 was also measured at the last word in the sentence. Under this condition, it was found that there was a reduced N400 for matching compared to mismatching pronouns, but only for participants who interpreted mismatching sentences to be ungrammatical.

In both models, condition was not found to be a significant predictor of surprisal (GRNN:$p=0.775$; JRNN: $p=0.112$). However, whether this is a successful replication of the responses of the participants who found the sentence to be grammatical (`Gramm.' in \Cref{fig:results}) or a failure to replicate the results of those who found the sentence ungrammatical (`Ungramm.' in \Cref{fig:results}) is unclear without further research, and thus this result is not discussed further in this paper.

\subsubsection{Semantic Anomaly}
In parallel to the pronoun stimuli, \citet{osterhout1995event} also compared N400 responses to sentences under the following experimental conditions: \textit{The boat sailed down the river and \textbf{sank}} (\textsc{control}; C) / \textit{\textbf{coughed}} (\textsc{semantically anomalous}; SA) \textit{during the storm}.

\paragraph{Target Words}
N400 amplitude was significantly lower in response to the experimentally manipulated \textsc{control} words compared to \textsc{semantically anomalous} words. This effect was replicated in the surprisals of both models ($p<0.001$ for all tests).

\paragraph{Sentence-Final Words}
The N400 and surprisals to sentence-final words followed the same pattern as target words ($p<0.001$ for all tests).

\subsection{Ainsworth-Darnell et al. (1998)}
\citet{ainsworth1998dissociating} investigate the difference in N400 amplitude in response to syntactic and semantic anomaly, operationalized in the following way: \textit{The chef entrusted the recipe \textbf{to relatives} before he left Italy} (\textsc{control}; C) / \textit{The chef entrusted the recipe \textbf{to carrots} before he left Italy} (\textsc{semantic anomaly}; \textsc{Sem}) /\textit{ The chef entrusted the recipe \textbf{relatives} before he left Italy} (\textsc{syntactic anomaly}; \textsc{Syn}) / \textit{The chef entrusted the recipe \textbf{carrots} before he left Italy} (\textsc{double anomaly}; DA). While previous research argued that the N400 does not respond to \textsc{syntactic anomaly}, they found that the \textsc{control} nouns elicited lower N400 amplitudes than nouns in other conditions, but they did not find a significant difference between the \textsc{syntactic anomaly} and \textsc{semantic anomaly} conditions or between the \textsc{semantic anomaly }and \textsc{double anomaly} conditions. \citet{ainsworth1998dissociating} do not report a test comparing the \textsc{syntactic anomaly} and \textsc{double anomaly} conditions, but it should be noted that \textsc{syntactic anomaly} has a lower amplitude (based on the graphs) than \textsc{semantic anomaly}, so an unreported significant difference between these should not be ruled out.

Experimental condition is a significant predictor of both GRNN and JRNN surprisal ($p<0.001$). For both models, the surprisal is lower for words in the \textsc{control} condition compared to other conditions ($p<0.001$ for all pairwise comparisons), and there is no significant difference between word in the \textsc{syntactic anomaly} and \textsc{semantic anomaly} conditions (GRNN: $p=0.274$; JRNN: $p=0.056$). The surprisals of the two models differ in that while \textsc{double anomaly} words differ from \textsc{semantic anomaly} words in both models (GRNN: $p<0.001$; JRNN: $p<0.001$), they do not differ from the \textsc{syntactic anomaly} in GRNN surprisal but they do in JRNN surprisal (GRNN: $p=0.059$; JRNN: $p<0.001$). Based on these findings and inspection of the graphs in \citet{ainsworth1998dissociating}, it appears that syntactic anomaly of this kind has a larger relative effect on surprisal than N400 amplitude.

\subsection{Kim and Osterhout (2005): Experiment 1}
\paragraph{Experiment 1} \citet{kim_independence_2005} investigate whether words that violate the event-structure of the described event are still facilitated if they are related to the event being described. The stimuli were of the following form: \textit{ The murder had been \textbf{witnessed} in the dark} (\textsc{passive control}; PC) / \textit{The bystanders had been \textbf{witnessing} the crime} (\textsc{active control}; AC) / \textit{The murder had been \textbf{witnessing} by the three bystanders} (\textsc{attraction violation}; AV). General analysis found that condition only marginally predicted N400 amplitude, but pairwise comparison found one significant difference bwetween conditions: PC completions elicited lower-amplitude N400s than AC completions.

In both models, condition was a significant predictor of surprisal, and PCs elicited the lowest surprisals, followed ACs, followed by AVs ($p<0.001$ for all tests).

\subsection{Kim and Osterhout (2005): Experiment 2}
Experiment 2 added the \textsc{no-attraction violation} (NV) condition to the study, which is exemplified by the following sentence: \textit{The unpleasant cough syrup was \textbf{witnessing} in the dark}. These were compared to results of the PC and AV conditions in Experiment 1. There was a significant main effect of condition, with PCs and AVs eliciting significantly lower-amplitude N400s than NVs.

Condition was a significant predictor the surprisals of both RNNs, with PCs eliciting a lower surprisal than AVs, followed by NVs with the highest surprisals ($p<0.001$ for all tests).

\section{General Discussion}We compared human N400 responses with surprisal in two RNN-LMs presented with the same stimuli, in the interest of determining the extent to which exposure to linguistic input alone can account for this particular component of human language processing. The results confirmed previous findings that surprisal is generally a good predictor of N400 amplitude, while also clearly demonstrating limitations of the models at capturing the human behavior. 

\subsection{Successful Predictions}
The models effectively predicted certain kinds of contrast that the N400 is sensitive to.  

\paragraph{Cloze}The surprisals of both models for the \citet{kutas1993company} and \citet{ito2016predicting} studies show that the surprisal of a language model is sensitive to cloze probability in the same direction as N400 amplitude---higher-cloze words elicit lower N400 amplitudes than lower-cloze words, and the same is true of surprisal. 

\paragraph{Relatedness} The results of the \citet{kutas1993company} and \citet{ito2016predicting} experiments also show that surprisal matches N400 amplitude in that words that are related to the highest-cloze completion in terms of semantics, but not form, elicit a lower surprisal than semantically unrelated words, even controlling for these words' cloze.

\paragraph{Semantic typicality} The surprisals of both models to the stimuli from \citeauthor{urbach_quantifiers_2010}'s \citeyearpar{urbach_quantifiers_2010} three experiments demonstrate that the surprisal of a language model patterns in the same way as N400 amplitude in that more typical words (in a given context) elicit a lower surprisal than atypical words in the same context.

\paragraph{Semantic anomaly} While the results are framed in the opposite direction in the original studies, the results from the Anomaly stimuli from \citet{osterhout1995event} and Experiment 1 of \citet{ainsworth1998dissociating} show that, all else being equal, completions that are not semantically anomalous (labeled `controls' in these experiments) elicit a lower surprisal from language models than  semantically anomalous completions, which is the result reported for N400 amplitude in the original studies.

\paragraph{Event structure violations} The results for Experiment 2 of \citet{kim_independence_2005} show that both surprisal and N400 amplitude are reduced when a word is in line with event-structure norms, compared to a word that is not and is semantically unrelated to the preceding context.

\subsection{Limitations and further directions}
At the same time, there are areas where the predictive capabilities of the models are limited.

\paragraph{Quantifiers} While the surprisal of the models matched the significant differences in Experiments 2 and 3 of \citet{urbach_quantifiers_2010} based on typicality overall, it did not replicate the finding that N400 amplitude was less reduced for \textsc{typical} nouns when they appeared with \textsc{few} or \textsc{rarely} quantifiers. Thus, it may be the case that some more explicit (or at least more specific) representation of quantification is involved in the neurocognitive processes underlying the N400 than can be modeled by surprisal alone.

\paragraph{Event structure violations} Overall, the surprisal of both models is more sensitive to morphosyntactic or event structure violations than N400 amplitude is \citep[for a discussion on the extent to which these can be considered separate in the context of ERPs, see][]{kuperberg_separate_2016}. For the stimuli from both \citet{kim_independence_2005} experiments, despite the \textsc{attraction violation} stimuli eliciting both a significantly reduced N400 amplitude and surprisal compared to the \textsc{no-attraction violation} stimuli, surprisal remained significantly higher for \textsc{attraction violation} stimuli than either of the control stimuli, which is not the case with N400 amplitude. Thus, by contrast with the case of quantifiers discussed above \citep{urbach_quantifiers_2010}, which seems to require a more detailed semantic representation, shallower or broader semantic representation might be needed to capture responses to the kinds of stimuli presented in \citet{kim_independence_2005}. If the goal is to improve the extent to which models capture human behavior, then there might be ways to accomplish this. \citet{frank2017word}, for example, use cosine distance between the sum of the vectors of all the preceding words in the sentence and the target word to predict the BOLD response (using fMRI) in N400 areas. Given the collateral facilitation of words semantically related to the highest-cloze completions of sentences, it is not unreasonable to assume that a similar process of spreading activation may occur for the preceding as well as the predicted upcoming word in the sentence. One way to implement this could be to weight the RNN model's predictions of the next word by each word's similarity to a general sentence-vector such as that used by \citet{frank2017word} before the probabilities are transformed into surprisal\footnote{See \citeauthor{kuperberg_separate_2016}'s \citeyearpar{kuperberg_separate_2016} discussion on bag-of-word approaches to the N400.}.

\paragraph{Morphosyntactic Anomaly} While there has been some discussion about the extent to which event structure violation and morphosyntactic anomalies can be considered separate in the context of ERPs \citep[see, e.g.][]{kuperberg_separate_2016}, there are clear cases where the surprisal of the language models appear to be more sensitive to morphosyntactic anomaly than N400 amplitude is. This can be seen in humans in the results of Experiment 1 of \citet{ainsworth1998dissociating}, where words that exhibit either semantic or syntactic anomalies elicit equally reduced surprisal. By contrast, the models predict grammatical continuations to a sentence over ungrammatical ones. This leads to lower surprisals for semantically anomalous words that are syntactically acceptable than those that are both syntactically and semantically anomalous. This difference between humans and the models supports the idea that there needs to be some way to weight predictions by semantic relatedness to the preceding context. 

\section{Conclusions}
Previous work has found that surprisal is a good predictor of N400 amplitude overall. Comparisons of surprisal in RNN-LMs to human N400 responses to the same input sentences showed for the first time that suprisal manages to account for a wide range of phenomena found in human N400 experiments. But at the same time, there are linguistic phenomena where it overpredicts, and others where it underpredicts a significant difference in the human N400 response. From the perspective of human language processing, this suggests that the activation of semantic and lexical features indexed by the N400 cannot be entirely captured by exposure to linguistic input alone. Specifically, quantification, aspects of event structure, and morphosyntactic anomalies seem to require some other learning architecture than the bottom-up statistical learning represented by standard recurrent neural networks. From the perspective of model-building, in order to improve a language-model based cognitive model of the N400, we need to allow for the addition of more shallow semantic processing  (independent of syntax and event structure) such as an implementation of spreading activation.

\bibliographystyle{acl_natbib}
\bibliography{anthology,emnlp2020}

\end{document}